%% file: acl_latex.tex
\pdfoutput=1

\documentclass[11pt]{article}

\usepackage{acl}
\usepackage{hyperref}
\usepackage{times}
\usepackage{latexsym}
\usepackage{multirow}
\usepackage{booktabs}
\usepackage{microtype}
\usepackage{multirow}
\usepackage{latexsym}
\usepackage{booktabs}
\usepackage{courier}
\usepackage{amsmath}
\usepackage{booktabs} 
\usepackage{siunitx} 
\usepackage{subfig}
\usepackage{placeins}
\usepackage{tikz}
\usepackage{hyperref}

\usepackage[T1]{fontenc}

\usepackage[utf8]{inputenc}

\usepackage{microtype}

\usepackage{inconsolata}
\usepackage{soul}

\title{IITK at SemEval-2024 Task 2: Exploring the Capabilities of LLMs for Safe Biomedical Natural Language Inference for Clinical Trials
}

\author{Shreyasi Mandal \qquad Ashutosh Modi \\
Indian Institute of Technology, Kanpur (IIT Kanpur)\\
\texttt{\{shreyansi, ashutoshm\}@cse.iitk.ac.in} 
}

\begin{document}
\maketitle

\input{sections/abstract}
\input{sections/introduction}

\input{sections/background}

\input{sections/system}
\input{sections/experiments}
\input{sections/results}
\input{sections/conclusion}

\section{Acknowledgments}
We would like to thank the anonymous reviewers for their careful reading of our manuscript and their insightful comments and constructive criticism. Their feedback has greatly helped us to improve the clarity and rigor of this work.

\bibliography{custom}

\vfill\null
    
\appendix
\input{sections/appendix}

\end{document}

%% file: sections/abstract.tex
\begin{abstract}
Large Language models (LLMs) have demonstrated state-of-the-art performance in various natural language processing (NLP) tasks across multiple domains, yet they are prone to shortcut learning and factual inconsistencies. This research investigates LLMs' robustness, consistency, and faithful reasoning when performing Natural Language Inference (NLI) on breast cancer Clinical Trial Reports (CTRs) in the context of SemEval 2024 Task 2: Safe Biomedical Natural Language Inference for Clinical Trials. We examine the reasoning capabilities of LLMs and their adeptness at logical problem-solving. A comparative analysis is conducted on pre-trained language models (PLMs), GPT-3.5, and Gemini Pro under zero-shot settings using Retrieval-Augmented Generation (RAG) framework, integrating various reasoning chains. The evaluation yields an F1 score of \textbf{0.69}, consistency of \textbf{0.71}, and a faithfulness score of \textbf{0.90} on the test dataset.
\end{abstract}

%% file: sections/introduction.tex
\section{Introduction} \label{sec:intro}
Clinical trials serve as essential endeavors to evaluate the effectiveness and safety of new medical treatments, playing a pivotal role in advancing experimental medicine. Clinical Trial Reports (CTRs) detail the methodologies and outcomes of these trials, serving as vital resources for healthcare professionals in designing and prescribing treatments. However, the sheer volume of CTRs (e.g., exceeding 400,000 and proliferating)  presents a challenge for comprehensive literature assessment when developing treatments \cite{bastian2010seventy}. Natural Language Inference (NLI) \cite{bowman-etal-2015-large} emerges as a promising avenue for large-scale interpretation and retrieval of medical evidence bridging recent findings to facilitate personalized care \cite{deyoung2020evidence, sutton2020overview}. The SemEval 2024 Task 2 on the Natural Language Inference for Clinical Trials (NLI4CT) \cite{jullien-etal-2024-semeval} revolves around annotating statements extracted from breast cancer CTRs\footnote{\normalsize{\href{https://clinicaltrials.gov/ct2/home}{https://clinicaltrials.gov/ct2/home}}} and determining the inference relation between these statements and corresponding sections of the CTRs, such as Eligibility criteria, Intervention, Results, and Adverse events. By systematically intervening in the statements, targeting numerical, vocabulary, syntax, and semantic reasoning, the task aims to investigate Large Language Models (LLM)s' consistency and faithful reasoning capabilities. 

In this paper, we experiment with Gemini Pro \cite{team2023gemini}, GPT-3.5 \cite{brown2020language}, Flan-T5 \cite{longpre2023flan} and several pre-trained language models (PLMs) trained on biomedical datasets, namely BioLinkBERT \cite{yasunaga-etal-2022-linkbert}, SciBERT \cite{beltagy-etal-2019-scibert}, ClinicalBERT \cite{huang2019clinicalbert}. We conducted zero-shot evaluations of Gemini Pro and GPT-3.5, employing Retrieval Augmented Generation (RAG) framework \cite{lewis2020retrieval} integrating Tree of Thoughts (ToT) reasoning \cite{yao2023tree} facilitating multiple reasoning paths. Our experiments involved applying various instruction templates to guide the generation process. These templates were refined through manual comparison of the labels within the training dataset against those generated by the models. The PLMs were fine-tuned on the provided training dataset, while the Flan-T5 model was assessed under zero-shot conditions.

\begin{figure*}[t]
    \centering
    \includegraphics[scale=0.5]{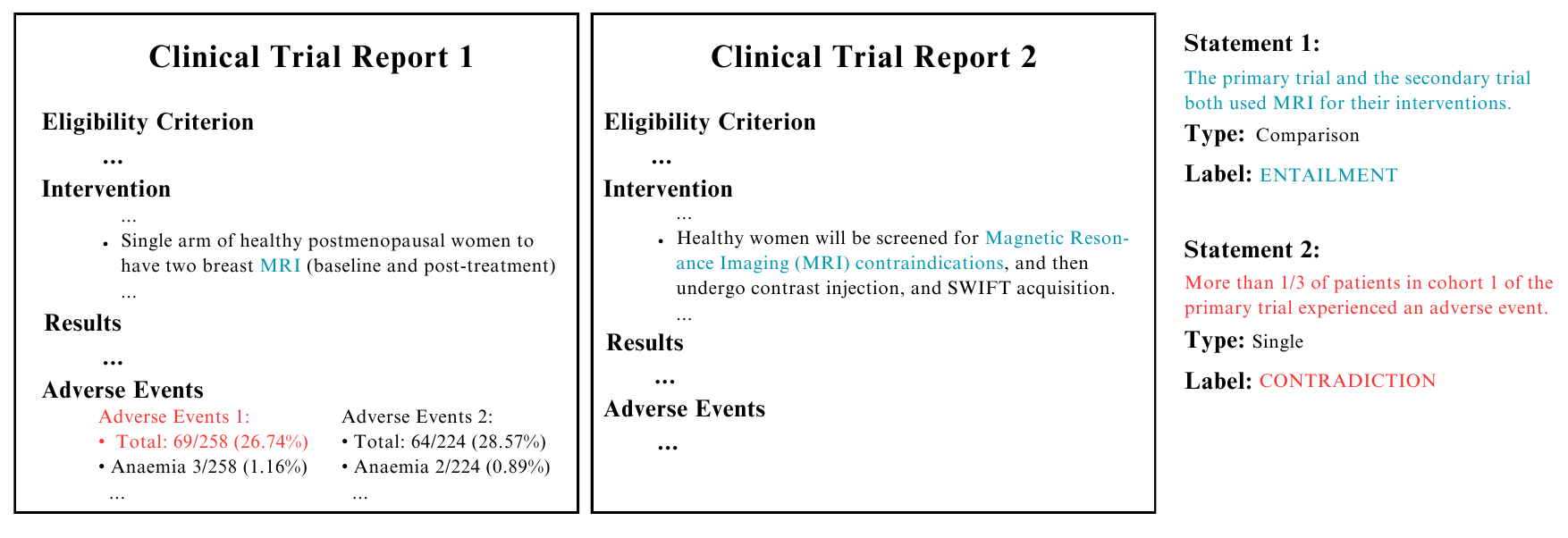}
    \caption{Examples of the dataset used in the NLI4CT task. Statement 1 compares the \textit{Intervention} section from two different clinical trial reports, while statement 2 is based on the \textit{Adverse Events} section of the first clinical trial report. The evaluation of the first statement requires textual inference skills, while the second requires numerical inference skills.}
    \label{fig:example-data}
\end{figure*}

Gemini Pro emerged as the top-performing model among all the experimented models, achieving an F1 score of \textbf{0.69}, with consistency and faithfulness scores of \textbf{0.71} and \textbf{0.90}, respectively, on the official test dataset. Notably, a comparative analysis between GPT-3.5 and Gemini Pro revealed shortcomings in GPT-3.5's performance, particularly in instances requiring numerical reasoning. For detailed examination of such instances, please refer to Appendix \ref{app-sec:examples}, where an example showcases GPT-3.5's accurate inference yet inadequate conclusion. The code to reproduce the experiments mentioned in this paper is publicly available.\footnote{\normalsize{\href{https://github.com/Exploration-Lab/IITK-SemEval-2024-Task-2-Clinical-NLI}{https://github.com/Exploration-Lab/IITK-SemEval-2024-Task-2-Clinical-NLI}}}




%% file: sections/background.tex
\section{Background} \label{sec:background}

\subsection{Related Work}
Pretrained Language Models (PLMs) and Large Language Models (LLMs) exhibit the potential to yield promising outcomes in the biomedical domain due to their ability to comprehend and process complex medical data effectively. BioLinkBERT \cite{yasunaga-etal-2022-linkbert}, pre-trained on PubMed\footnote{\normalsize{\href{https://pubmed.ncbi.nlm.nih.gov}{https://pubmed.ncbi.nlm.nih.gov}}}, utilizes hyperlinks within documents. It has attained state-of-the-art (SOTA) performance across a wide range of tasks and various medical NLP benchmarks, namely BLURB \cite{gu2021domain} and BioASQ \cite{nentidis2020results}. SciBERT \cite{beltagy-etal-2019-scibert} is trained on scientific publications from the biomedical domain in Semantic Scholar\footnote{\normalsize{\href{https://www.semanticscholar.org}{https://www.semanticscholar.org}}}. ClinicalBERT \cite{huang2019clinicalbert} is trained using clinical text data sourced from approximately 2 million clinical notes contained within the MIMIC-III database \cite{johnson2016mimic}. \citet{kanakarajan2023saama} employed a fine-tuned Flan-T5-xxl model with instruction tuning, achieving an F1 score of 0.834 on the SemEval 2023 Task 7 \cite{jullien-etal-2023-nli4ct,jullien-etal-2023-semeval}. \citet{zhou2023thifly} performed joint semantics encoding of the clinical statements followed by multi-granularity inference through sentence-level and token-level encoding, getting an F1 score of 0.856. Although these models have achieved high performance, there remains a need for further investigation into their application in vital areas such as real-world clinical trials.

\textbf{GPT-3.5}, developed by OpenAI\footnote{\href{https://openai.com}{https://openai.com}} and comprising 175 billion parameters, uses alternating dense and locally banded sparse attention patterns in the transformer layers \cite{child2019generating, wolf2020transformers}. The token size limit for GPT-3.5 (free tier) is 4,096. \textbf{Gemini Pro}, developed by Google DeepMind\footnote{\href{https://deepmind.google}{https://deepmind.google}} uses decoder-only transformers \cite{vaswani2017attention} and multi-query attention \cite{shazeer2019fast} with a context window length of 32,768 tokens.

\begin{table}[h]
\centering
\resizebox{\columnwidth}{!}{
\begin{tabular}{@{}lccc@{}}
\toprule
\textbf{Data} & \textbf{Number of Samples} & \multicolumn{2}{c}{\textbf{Labels}} \\
\cline{3-4}
& & Entailment & Contradiction\\
\midrule
train & 1700 & 850 & 850 \\
dev & 200 & 100 & 100 \\
test & 5500 & 1841 & 3659 \\
\bottomrule
\end{tabular}%
}
\caption{The number of samples in each subset of the data. The distribution of the labels between the train and the development set is even. Note: The test set labels were made public after the completion of the task.}
\label{table:data}
\end{table}

\begin{figure*}[t]
    \centering
    \includegraphics[scale=0.60]{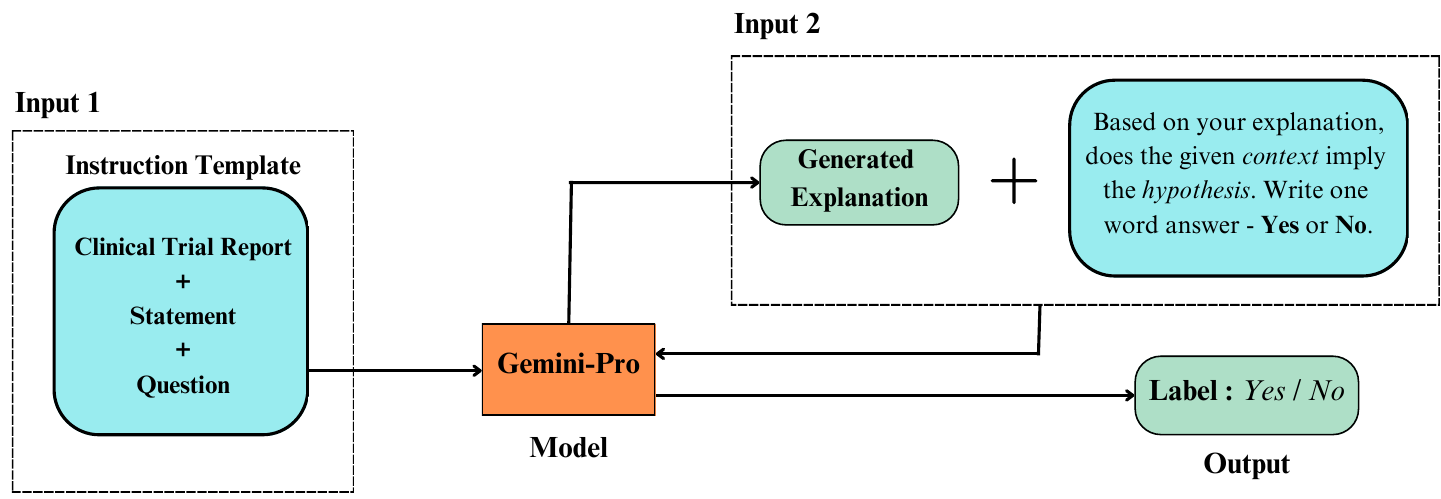}
    \caption{An overview of the proposed system architecture used for the NLI4CT Task}
    \label{fig:models}
\end{figure*}

\subsection{Task and Dataset Description} \label{subsec:data}
The NLI4CT task \cite{jullien-etal-2024-semeval} focuses on textual entailment based on a collection of breast cancer CTRs, statements, explanations and labels annotated by domain expert annotators. The CTRs are in English. The CTRs are segmented into four sections - eligibility criteria, intervention details, results, and adverse events. The statements, with an average length of 19.5 tokens, make claims about the information contained in one of the sections of a CTR or compare the same section from two different CTRs as seen in Figure \ref{fig:example-data}. The task involves determining the inference relation (\textit{entailment} or \textit{contradiction}) between CTR-statement pairs. The dataset consists of 999 Clinical Trial Reports (CTRs) and 7400 annotated statements, which are divided into train, development and test sets. Table \ref{table:data} provides statistics for the dataset.


%% file: sections/system.tex
\section{System Overview} \label{sec:system}



LLMs such as GPT-3 \cite{brown2020language} and Gemini Pro \cite{team2023gemini} have shown remarkable performances across various tasks. For the NLI4CT task, we have experimented with Gemini Pro, GPT-3.5, Flan-T5 \cite{longpre2023flan}, BioLinkBERT \cite{yasunaga-etal-2022-linkbert}, SciBERT \cite{beltagy-etal-2019-scibert}, ClinicalBERT \cite{huang2019clinicalbert} and ClinicalTrialBioBert-NLI4CT\footnote{\href{https://huggingface.co/domenicrosati/ClinicalTrialBioBert-NLI4CT}{https://huggingface.co/domenicrosati/ClinicalTrialBioBert-NLI4CT}}. The performance of the different models is shown in Figure \ref{fig:plm-llms-results}. Zero-shot evaluation was done on Gemini Pro and GPT-3.5, Flan-T5 was instruction fine-tuned following  \citet{kanakarajan2023saama}, and the rest of the models were trained on the given train and development dataset. Gemini Pro and GPT-3.5 were considered for further experimentation because of their superior performance.

The proposed system utilizes structured instruction templates and multi-turn conversation techniques to generate explanations and labels for the statements provided as input, as shown in Figure \ref{fig:models}.

Reasoning is an essential ability required by an LLM to solve complex problems \cite{qiao2022reasoning}. Tree of Thoughts (ToT) framework \cite{yao2023tree} and Chain-of-Thought (CoT) reasoning \cite{wei2022chain} is integrated into the models, facilitating multiple reasoning paths. 

\subsection{Reasoning Frameworks}
\begin{figure}[h!]
    \centering
    \includegraphics[width=0.40\textwidth]{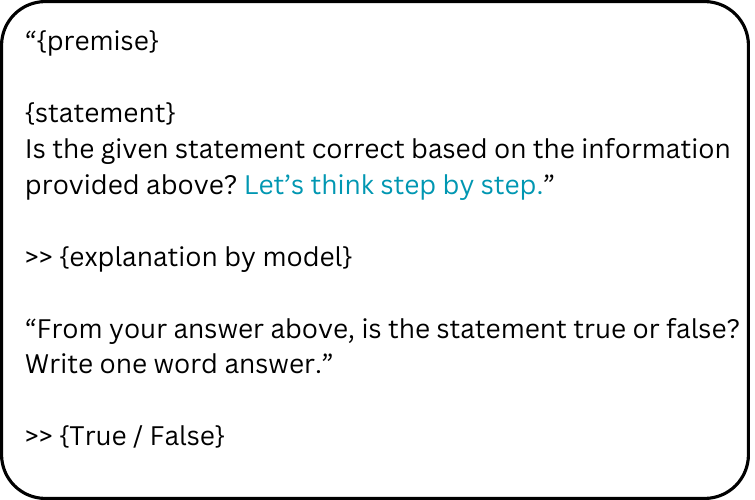}
    \caption{Instruction template for CoT prompting}
    \label{fig:cot}
\end{figure}

\begin{figure}[h!]
    \centering
    \includegraphics[width=0.5\textwidth]{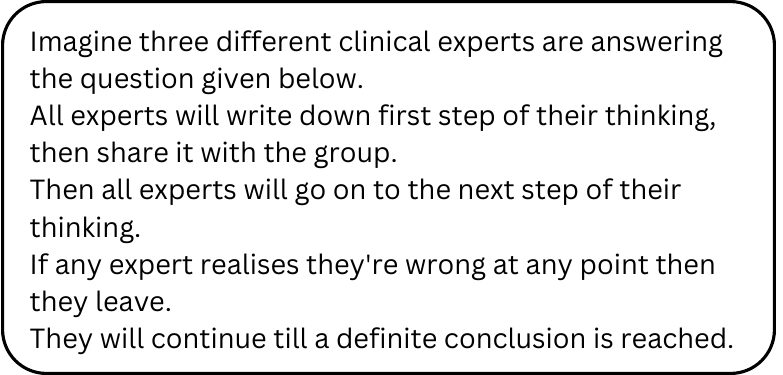}
    \caption{Prompt for Tree of Thought reasoning}
    \label{fig:prompt}
\end{figure}

\textbf{Chain-of-Thought (CoT)} prompting \cite{wei2022chain} has demonstrated promising results in improving the reasoning abilities of LLMs. To evaluate Gemini Pro and GPT-3.5, we used Zero-shot-CoT \cite{kojima2022large} prompt reasoning without requiring few-shot demonstrations. The phrase \textit{``Let’s think step by step''} is added after the instruction as shown in Figure \ref{fig:cot}.
\\
\textbf{Tree-of-Thought (ToT)} framework \cite{yao2023tree, long2023large} relies on trial and error method to solve complex reasoning tasks. It facilitates multi-round conversations and backtracking. Our system allows for three reasoning paths using the prompt shown in Figure \ref{fig:prompt}.\footnote{\href{https://github.com/dave1010/tree-of-thought-prompting}{https://github.com/dave1010/tree-of-thought-prompting}} 

For the evaluation of the model, the input to Gemini Pro and GPT-3.5 is constructed using an instruction template containing the appropriate prompt for ToT or CoT reasoning, data from the CTR which constitutes the premise and the statement or the hypothesis as shown in Figure \ref{fig:models}. A series of two questions is presented to the model to generate both the explanation and the corresponding label. Multi-turn conversation \cite{zhang2018modeling} is used to include the generated explanation as context for generating the final label. The explanation is also retained for further experimentations. The generated final label is converted as follows: \textit{\{"Yes": "Entailment", "No": "Contradiction"\}}. A comparison of the performance of GPT-3.5 and Gemini Pro after integrating CoT and ToT reasoning frameworks is shown in Figure \ref{fig:reas-comp}. 


    
    





%% file: sections/experiments.tex
\section{Experimental setup} \label{sec:experiments}

\subsection{Data Preprocessing}
As discussed in Section \ref{subsec:data}, the statements can make claims about the information contained in one of the sections of a CTR, which is then called a {``Single''} statement or compare the same section from two different CTRs, called a {``Comparison''} statement.
In {``Single''} statements, the term ``primary'' is employed to assert a claim. Evidence from the CTR is compiled into a unified text structure, exemplified as follows: \textit{``For the primary trial participants, \{primary evidences\}''}. In contrast, for {``Comparison''} statements, the term ``secondary'' accompanies ``primary''. The evidences are then compiled as: \textit{``For the primary trial participants, \{primary evidences\}. For the secondary trial participants, \{secondary evidences\}''}.

\subsection{Hyperparameter Tuning}
For Gemini Pro, the temperature of the model is set to 0.7 and the safety settings are set to "BLOCK\_NONE". For GPT-3.5, the models \textit{"gpt-3.5-turbo-0613"} and \textit{"gpt-3.5-turbo-1106"} are used for experimentation among which \textit{"gpt-3.5-turbo-0613"} performs considerably better. The temperature of the model is set to 0.6.

\begin{figure*}[t]
    \centering
    \includegraphics[scale=0.57]{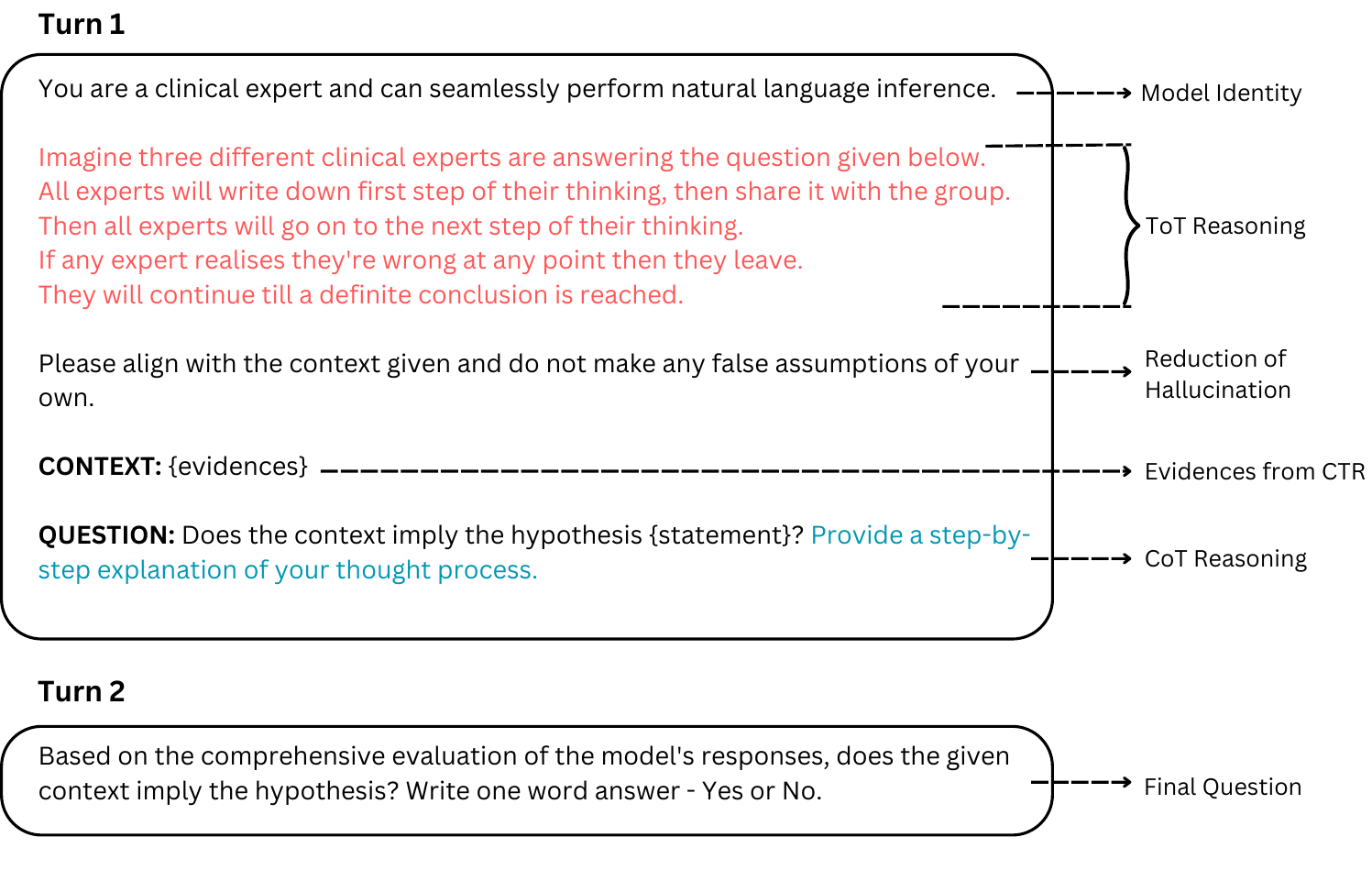}
    \caption{Final Instruction Template}
    \label{fig:final-it}
\end{figure*}

\subsection{Prompt Engineering}
The system was experimented with several prompts to improve its performance. The explanations generated by the model were examined manually to identify instances where the solution deviated from the correct path. The prompt \textit{``You are a clinical expert and can seamlessly perform natural language inference''} was introduced to give the model an identity. Additionally, rules were enforced to confine the model's output within the provided context and to prevent hallucinations, achieved through the prompt: \textit{``Please align with the context given and do not make any false assumptions of your own.''} Furthermore, to integrate CoT reasoning within the ToT framework, the prompt \textit{''Provide a step-by-step explanation of your thought process''} was introduced. The final instruction template is shown in Figure \ref{fig:final-it}.

Several experiments were conducted to assess the model's performance on extracting the labels \textit{"Entailment"} or \textit{"Contradiction"} in the second question of the multi-turn conversation. The F1 scores for various prompts on the development set are presented in Table \ref{table:prompts}. Ultimately, Prompt 4 demonstrated the best performance and was chosen for the final pipeline.
\begin{table}[h!]
\centering
\resizebox{\columnwidth}{!}{
\begin{tabular}{@{}lc@{}}
\toprule
\textbf{Prompt} & \textbf{F1 score} \\
\midrule
\textit{Based on the comprehensive evaluation}\\\textit{of the model's responses, is the given}\\\textit{hypothesis deemed to be true or false?}\\\textit{Write one word answer.} & 0.689 \\
\midrule
\textit{Does this imply that the given hypothesis}\\\textit{is supporting the report or not? Give one}\\\textit{word answer (Yes / No).} & 0.667 \\
\midrule
\textit{From your answer above, is the statement}\\\textit{true or false? Write one word answer.} & 0.656 \\
\midrule
\textit{Based on your explanation, does the given}\\\textit{context imply the hypothesis. Write one}\\\textit{word answer.} & \textbf{0.723} \\

\bottomrule
\end{tabular}%
}
\caption{Performance of the model on the \textit{dev} data for different prompts for extracting the labels}
\label{table:prompts}
\end{table}

\subsection{Evaluation Metrics}
The NLI4CT task \cite{jullien-etal-2024-semeval} is evaluated on the basis of three metrics - F1 score, consistency and faithfulness. Faithfulness measures the accuracy of the system's predictions by evaluating its ability to predict outcomes for altered inputs correctly. If the model correctly adjusts its predictions in response to semantic alterations, it demonstrates higher faithfulness. On the other hand, consistency evaluates the model's ability to provide consistent predictions for semantically equivalent inputs.






%% file: sections/results.tex
\section{Results} \label{sec:results}
The zero-shot evaluation of Gemini Pro yielded an F1 score of \textbf{0.69}, with a consistency of \textbf{0.71} and a faithfulness score of \textbf{0.90} on the official test dataset. Our system achieved a fifth-place ranking based on the faithfulness score, a sixteenth-place ranking based on the consistency score, and a twenty-first-place ranking based on the F1 score. Gemini Pro outperforms GPT-3.5 with an improvement in F1 score by $+$1.9\%, while maintaining almost similar consistency score. Additionally, the faithfulness score of Gemini Pro improves by $+$3.5\% compared to GPT-3.5, as illustrated in Table \ref{table:result}.
\begin{table}[h!]
\centering
\resizebox{\columnwidth}{!}{
\begin{tabular}{@{}lccc@{}}
\toprule
\textbf{Model} & \textbf{Base F1} & \textbf{Consistency} & \textbf{Faithfulness} \\
\midrule
Gemini Pro & \textbf{0.691} & 0.712 & \textbf{0.901} \\
GPT-3.5 & 0.672 & \textbf{0.713} & 0.866 \\
\bottomrule
\end{tabular}%
}
\caption{Results on the \textit{test} data using Gemini Pro and GPT-3.5}
\label{table:result}
\end{table}

The system utilizing Gemini Pro attained an F1 score of 0.72, while GPT-3.5 achieved an F1 score of 0.68 on the training dataset. Manual examination of the model-generated explanations and a comparison of the generated labels with the original labels was conducted to refine the prompts and enhance the model's responses.

\begin{figure}[h!]
    \centering
    \includegraphics[width=0.47\textwidth]{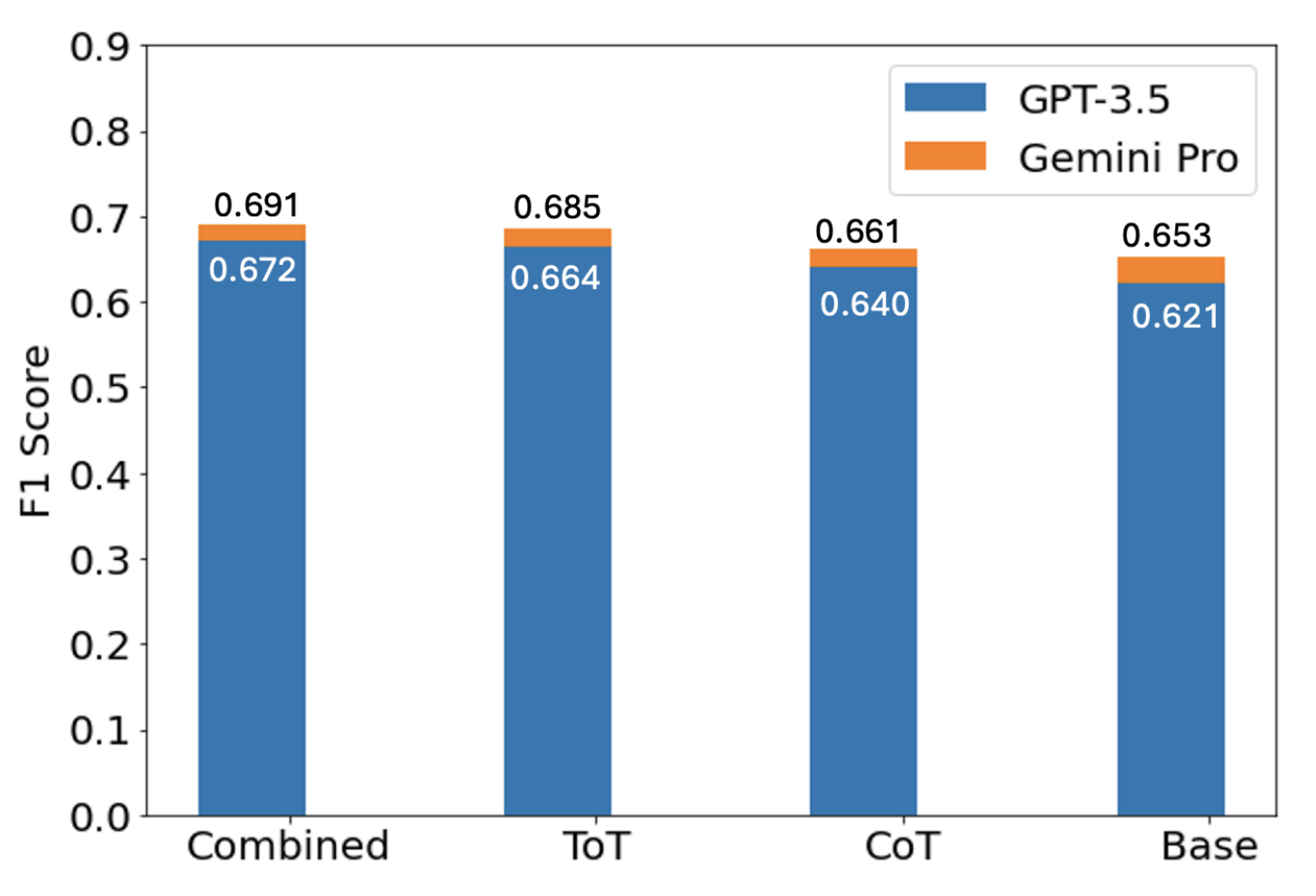}
    \caption{Comparison of the performance of Gemini Pro and GPT-3.5 without the integration of any reasoning framework, with CoT reasoning, with ToT reasoning and with both the reasoning frameworks combined.}
    \label{fig:reas-comp}
\end{figure}

\begin{figure}[h!]
    \centering
    \includegraphics[scale=0.25]{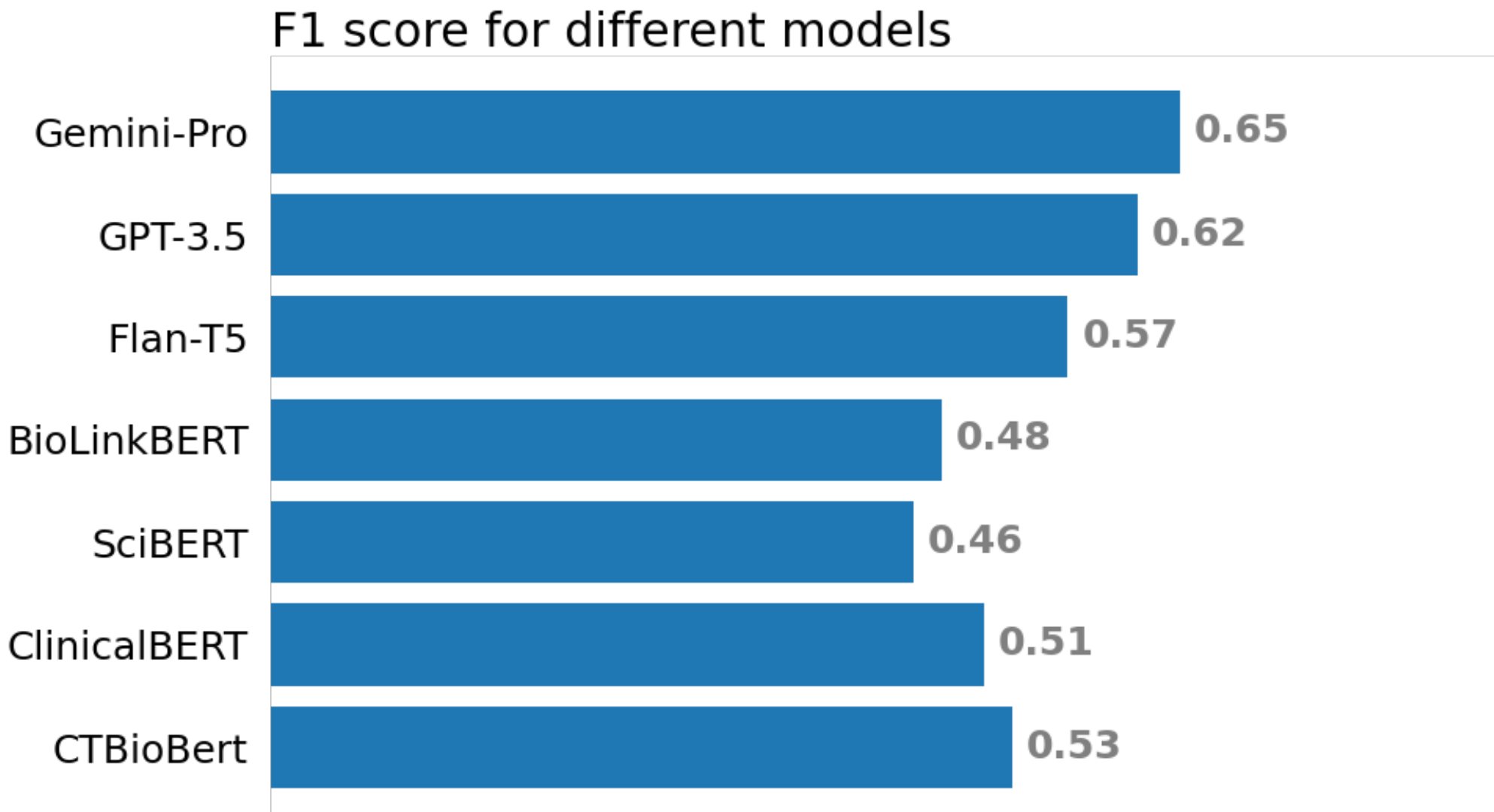}
    \caption{Performance (F1 Score) of the different experimented models. Note: CTBioBert represents the model ClinicalTrialBioBert-NLI4CT.}
    \label{fig:plm-llms-results}
\end{figure}

As depicted in Figure \ref{fig:reas-comp}, the integration of CoT reasoning led to an increase in performance for Gemini Pro and GPT-3.5 by 0.8\% and 1.9\%, respectively. Furthermore, upon integrating the ToT reasoning framework, the performance improved by 3.2\% and 4.3\%, respectively. When both ToT and CoT reasoning were integrated, the models showed an increase in performance by \textbf{3.8\%} and \textbf{5.1\%}, respectively, compared to the baseline model.
\\
Figure \ref{fig:plm-llms-results} compares the performance of Gemini Pro and GPT-3.5, both without reasoning frameworks, with Flan-T5 and other experimented PLMs. Gemini Pro achieved the highest F1 score of 0.65, followed closely by GPT-3.5 with an F1 score of 0.62. Flan-T5 performed moderately with an F1 score of 0.57, while BioLinkBERT, SciBERT, ClinicalBERT, and CTBioBERT displayed lower F1 scores ranging from 0.46 to 0.53.

A comparative analysis between GPT-3.5 and Gemini Pro highlighted GPT-3.5's shortcomings in tasks requiring logical reasoning. Appendix A presents the example responses for both the models. The appendix further analyzes potential reasoning errors made by GPT-3.5 and Gemini Pro.





%% file: sections/conclusion.tex
\section{Conclusion} \label{sec:conclusion}
This paper presents an evaluation of several pre-trained language models (PLMs), and GPT-3.5, Gemini Pro, under zero-shot conditions. Our analysis focuses on assessing the reasoning capabilities of GPT-3.5 and Gemini Pro and their adeptness at logical problem-solving. In the NLI4CT task, we achieved an F1 score of 0.691, consistency of 0.71, and faithfulness of 0.90. Additionally, our findings underscore that prompt engineering is crucial for large language models (LLMs). We have made our instruction templates and code publicly available to facilitate reproducibility.

%% file: sections/appendix.tex
\section*{Appendix}

\section{Example Responses} \label{app-sec:examples}

\begin{figure}[h!]
    \centering
    \includegraphics[width=0.5\textwidth]{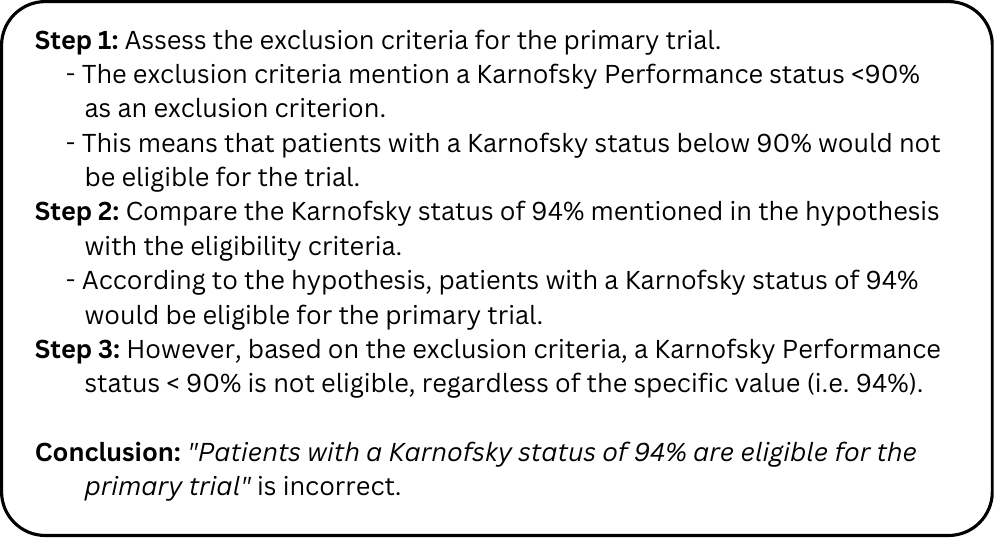}
    \caption{Example of an incorrect response by GPT-3.5}
    \label{fig:gpt}
\end{figure}
\begin{figure}[h!]
    \centering
    \includegraphics[width=0.5\textwidth]{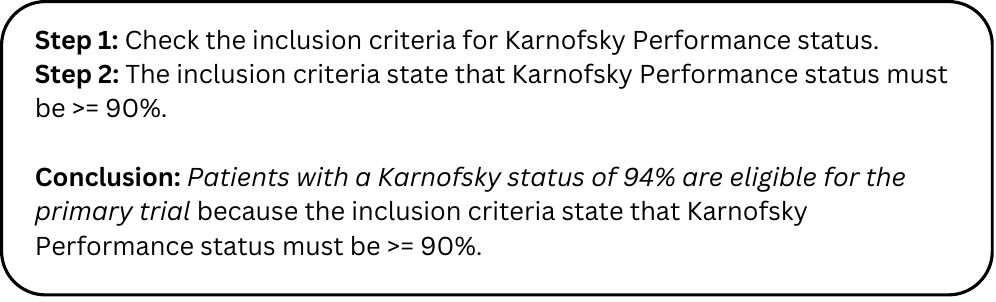}
    \caption{Response by Gemini Pro for the same statement}
    \label{fig:gemini}
\end{figure}

As illustrated in Figure \ref{fig:gpt}, GPT-3.5 exhibits limitations in numerical reasoning tasks. Although steps 1 and 2 of the example are executed correctly, GPT-3.5 starts hallucinating in step 3, leading to illogical reasoning and ultimately yielding an incorrect conclusion that the statement is incorrect. When the same input is given to Gemini Pro, it correctly concludes that the statement entails the CTR as shown in Figure \ref{fig:gemini}.
\vspace{-2mm}
\vspace{-3mm}

\begin{figure}[h!]
    \centering
    \includegraphics[width=0.5\textwidth]{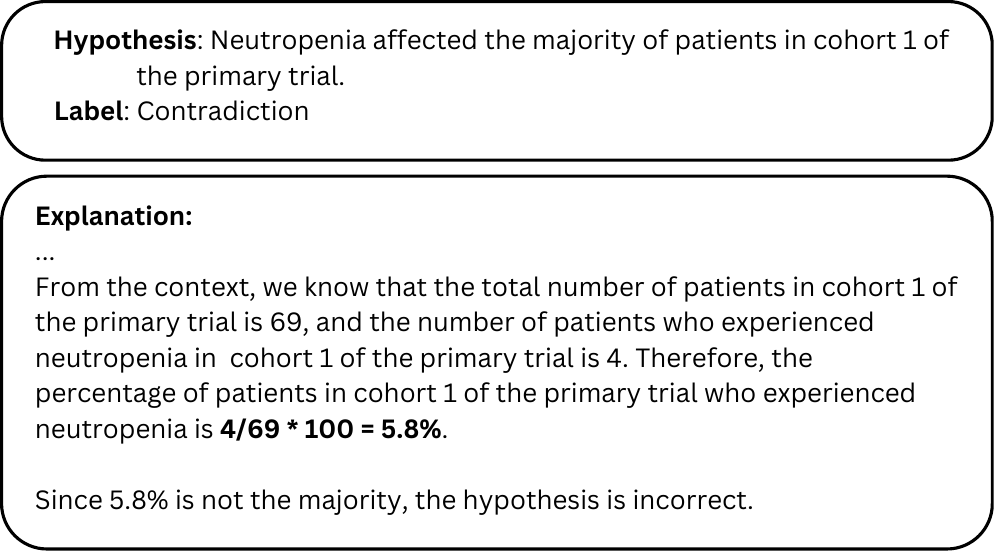}
    \caption{An example response by Gemini Pro showcasing its mathematical reasoning ability.}
    \label{fig:gemini-math}
\end{figure}

Figure \ref{fig:gemini-math} showcases an excellent example of Gemini Pro's mathematical reasoning. 
